%% file: main.tex
\documentclass[10pt,twocolumn,letterpaper]{article}

\usepackage{cvpr}              %

\input{preamble}

\definecolor{cvprblue}{rgb}{0.21,0.49,0.74}
\usepackage[pagebackref,breaklinks,colorlinks,citecolor=cvprblue]{hyperref}

\usepackage{graphicx}
\usepackage{amsmath}
\usepackage{amssymb}
\usepackage{xfrac}
\usepackage{color}
\usepackage{colortbl}
\usepackage{enumitem}
\usepackage[dvipsnames]{xcolor}

\usepackage{sidecap}
\usepackage{wrapfig}
\usepackage{lipsum} 
\usepackage{textcmds}
\usepackage{graphicx}
\usepackage{multirow}
\usepackage{booktabs}
\usepackage{tabularx}
\usepackage{colortbl}
\usepackage{xcolor}
\usepackage{hhline}
\usepackage{lipsum}
\usepackage{etoolbox}
\usepackage{enumerate}
\usepackage{diagbox}
\usepackage{algorithm}
\usepackage{algpseudocode}
\usepackage[accsupp]{axessibility}
\usepackage{pifont}
\usepackage{capt-of,etoolbox}
\definecolor{Gray}{gray}{0.90}
\definecolor{LightCyan}{rgb}{0.82,0.82,1}
\definecolor{tabhighlight}{HTML}{e5e5e5}

\def\etal{\emph{et al.}\xspace}

\def\eg{\emph{e.g.}\xspace}

\def\paperID{} %
\def\confName{MMFM}
\def\confYear{2024}

\title{X-MIC: Cross-Modal Instance Conditioning \\for Egocentric Action Generalization}

\author{Anna Kukleva$^{1,2*}$ \hspace*{.3in} Fadime Sener$^1$ \hspace*{.3in} Edoardo Remelli$^1$ \hspace*{.3in} Bugra Tekin$^1$ \hspace*{.3in} Eric Sauser$^1$ \\ Bernt Schiele$^2$ \hspace*{.3in} Shugao Ma$^1$\\
$^1$Meta Reality Labs; $^2$Max Planck Institute for Informatics, Saarland Informatics Campus  \\
{\tt\small \{annakukleva, famesener\}@meta.com}
}

\begin{document}
\maketitle
\input{sec/0_abstract}    
\input{sec/1_intro}
\input{sec/2_related_work}
\input{sec/3_method}
\input{sec/4_results}

\input{sec/5_conclusion}

{
    \small
    \bibliographystyle{ieeenat_fullname}
    \bibliography{main}
} 

\input{sec/X_suppl}

\end{document}

%% file: preamble.tex
\usepackage[dvipsnames]{xcolor}
\usepackage{soul}

\newcommand{\our}[0]{X-MIC\xspace}

\newcommand{\myparagraph}[1]{\vspace{2pt}\noindent{\textbf{#1}}}

%% file: sec/0_abstract.tex
\begin{abstract} 

Lately, there has been growing interest in adapting vision-language models~(VLMs) to image and third-person video classification due to their success in zero-shot recognition. However, the adaptation of these models to egocentric videos has been largely unexplored. To address this gap, we propose a simple yet effective cross-modal adaptation framework, which we call X-MIC. Using a video adapter, our pipeline learns to align frozen text embeddings to each egocentric video directly in the shared embedding space. Our novel adapter architecture retains and improves generalization of the pre-trained VLMs by disentangling learnable temporal modeling and frozen visual encoder. This results in an enhanced alignment of text embeddings to each egocentric video, leading to a significant improvement in cross-dataset generalization. We evaluate our approach on the Epic-Kitchens, Ego4D, and EGTEA datasets for fine-grained cross-dataset action generalization, demonstrating the effectiveness of our method.\footnote{\url{https://github.com/annusha/xmic} \\ \hspace*{.1in} *work is done during internship at Meta}
 
\end{abstract}

%% file: sec/1_intro.tex
\section{Introduction}
\label{sec:intro}
 
\begin{figure}[!t]
\begin{center}
\includegraphics[width=0.90\linewidth]{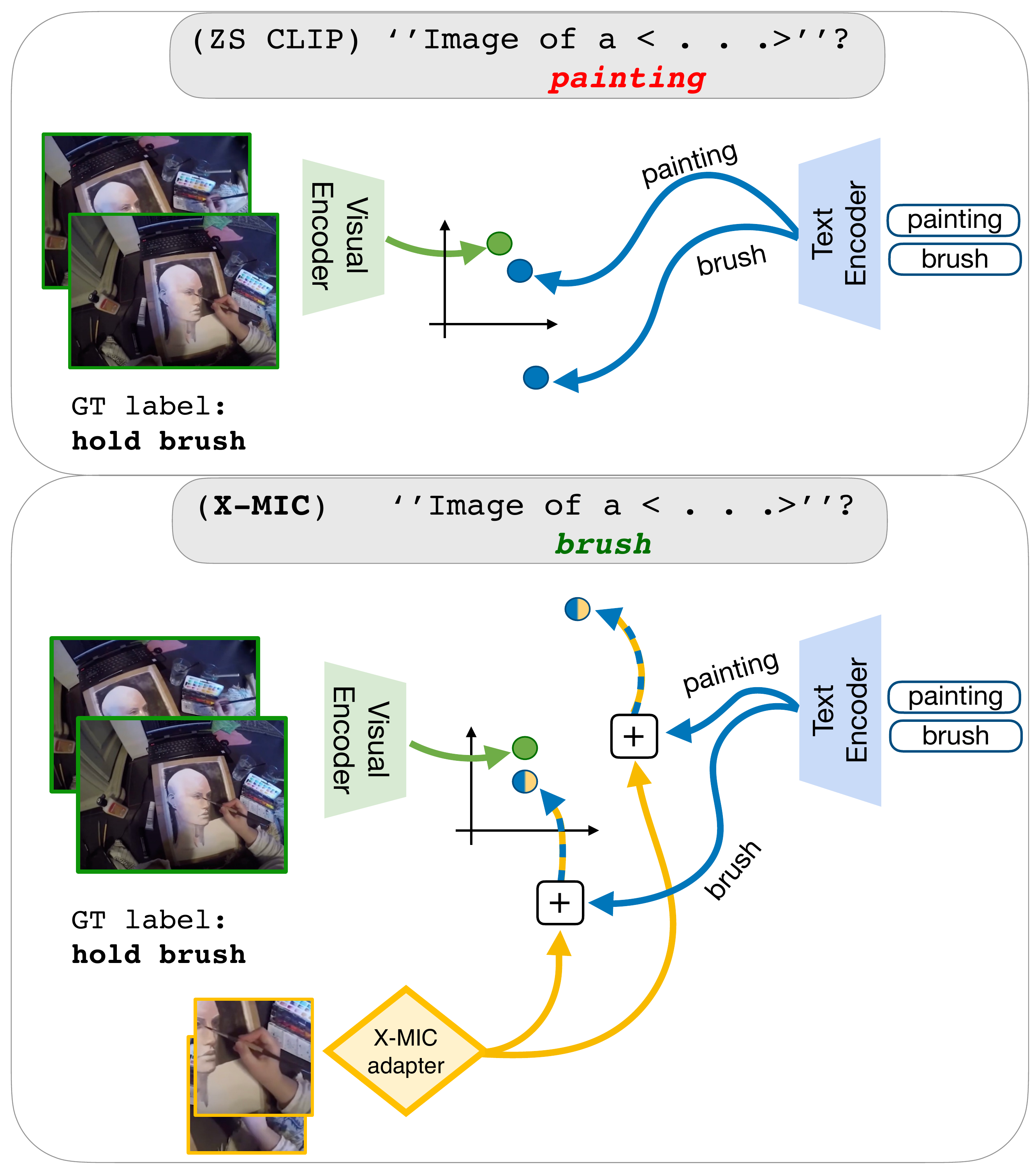}
\end{center} 
\vspace{-5mm}
\caption{
\textbf{Egocentric video classification with VL models.}\\ \textbf{Top:} Standard zero-shot CLIP. As the dominant object in the scene is painting, the model predicts class ``painting'' while the object of interest is ``brush''. \textbf{Bottom:} CLIP model with our X-MIC adaptation directly in the shared VL embedding. X-MIC vectors adapt focus of the CLIP model to the hand area, guiding text modality to capture egocentric domain-specific information.
}
\vspace{-6mm}
\label{fig:teaser} 
\end{figure}

Egocentric action recognition has recently become a popular research topic due to the rising interest in augmented reality and robotics. Recently, two large-scale egocentric datasets Epic-Kitchens~\cite{Damen2022RESCALING} and Ego4D~\cite{Ego4D2022CVPR}, capturing the daily activities of users have been introduced. While there is a growing interest in studying action recognition on egocentric datasets, evaluations primarily occur within the same dataset; lacking cross-dataset evaluations that is crucial for real-world deployment of recognition models. Testing models on different datasets presents several challenges, such as encountering unfamiliar environments, different users, and previously unseen objects and their corresponding actions, all of which can significantly impact performance. Recently, vision-language models~\cite{jia2021scaling,radford2021learning,zhang2022contrastive} such as CLIP~\cite{radford2021learning} have demonstrated remarkable performance across diverse third-persons datasets like Kinetics-600~\cite{kay2017kinetics} and ImageNet~\cite{imagenet_cvpr09}, showcasing their ability to generalize effectively and achieving zero-shot performance of 59.8\% and 76.2\%, respectively. However, their zero-shot performance drops significantly when applied to egocentric datasets like Epic-Kitchens, with noun and verb recognition reaching only 8.8\% and 5.9\%, respectively; highlighting the domain gap between third-person and egocentric data. 
 
CLIP's zero-shot generalization to new datasets leverages learning a shared embedding space for text and visual modalities. To enhance generalization to new domains, a prominent research direction~\cite{zhou2022learning} explores adapting the text encoder by appending trainable \textit{prompt} tokens to class tokens, modifying the class-text input from ``a photo of an apple'' to ``$<$learnable prompt$>$ apple''. As an alternative approach, recent work has proposed to train feature \textit{adapters} on both the visual and textual domains~\cite{chen2022vision,gao2023clip}, drawing insights from the NLP works~\cite{houlsby2019parameter,stickland2019bert}. Despite their promising results, these methods overlook the inherent characteristics of the egocentric video domain. To overcome this, we propose a simple yet effective adapter architecture, injecting egocentric video-specific knowledge into a frozen VL embedding space, depicted in Fig.~\ref{fig:teaser}. Our method transforms each video through an adapter into a vector for \underline{cross}-\underline{m}odal \underline{i}nstance \underline{c}onditioning of text — referred to as \our-vector. Our cross-modal adaptation performed directly in the embedding space results in significantly improved efficiency during training and testing. Moreover, our new adapter module disentangles frozen visual encoder from the visual temporal modeling through cross-modal adaptation. Each \our-vector is video-specific, therefore, allowing us to align any frozen text to each input video individually. Finally, to align the text embedding to the video, we simply add the \our-vector to the text embedding vectors.
 
We extensively evaluate our approach on Epic-Kitchens~\cite{Damen2022RESCALING}, Ego4D~\cite{Ego4D2022CVPR} and EGTEA~\cite{li2018egtea} datasets, demonstrating superior generalization compared to SOTA VL-adaptation methods. 

\noindent Our contributions can thus be summarized as: 
\begin{itemize}
\item 
addressing the task of egocentric cross-dataset and zero-shot action recognition with VLMs that is designed for real-world applications, e.g. AR, addressing the impracticality of collecting data from every new environment,
\item 
a simple yet effective framework, referred to as \our, for cross-modal adaption of VL models directly in the pre-trained VL embedding space; our module disentangles temporal modeling from the frozen visual encoder,
\item 
a new egocentric spatial-temporal attention module enhances information around hands, thereby improving egocentric action recognition performance, 
\item 
thorough comparisons with respect to image and video state-of-the-art VL adaptation methods which demonstrate the effectiveness of our approach. 
\end{itemize}

%% file: sec/2_related_work.tex
\section{Related Work}
\label{sec:rel_work}
 
\myparagraph{Egocentric Action Generalization.} 
While egocentric vision gained attention with datasets like Epic-Kitchens~\cite{Damen2022RESCALING} and Ego4D~\cite{Ego4D2022CVPR}, current state-of-the-art~\cite{sener2021technical,zhao2022real,xiong2022m,patrick2021keeping,wu2022memvit,kazakos2019epic,xiao2020audiovisual,zhukov2019cross,sener2022assembly101} primarily focus on intra-dataset evaluation, which limits their applicability to real-world scenarios. Several methods fine-tuned CLIP on egocentric datasets~\cite{zhao2023lavila,lin2022egocentric,pramanick2023egovlpv2}, yet generalization on fine-grained verbs and nouns recognition remains underexplored. Our work comprehensively investigates both intra-dataset and inter-dataset generalization on both verbs and nouns. 

\myparagraph{Prompt Learning and Adapters.} 
Prompt learning in NLP~\cite{zhong2021factual,shin2020autoprompt,jiang2020can,gao2020making,lester2021power} adapts frozen text models by appending task-specific information. Extending this to image recognition, CoOp~\cite{zhou2022learning} learns appendable vectors in the text token space. CoCoOp~\cite{zhou2022conditional} introduces image-conditioned prompt learning, boosting performance but with high computational costs. MaPLe~\cite{khattak2023maple} leverages shared deep prompts for text and visual encoders, while PromptSRC~\cite{rasheed2023fine} suggests regularizing constraints for frozen encoders. Chen~\etal~\cite{chen2022adaptformer} found that prompting boosts model transferability in tasks with fewer number of visual tokens, like image classification, but has limited impact in tasks with more tokens, such as video understanding. Thus, an alternative research direction explores adapting vision-language models with feature adapters~\cite{houlsby2019parameter}. Clip-adapter~\cite{gao2023clip} learns new features through an additional bottleneck layer and blends them with original pre-trained features in a residual style. Our approach falls under the adapter category. Unlike previous visual adapters, we introduce cross-modal instance conditioning specifically designed for egocentric video recognition. 

\myparagraph{Adapting VLMs to Videos. } Recent advancements in prompt learning extend to third-person videos. A5/A6~\cite{ju2022prompting} introduces a temporal module atop visual encoder, keeping both encoders frozen. EVL~\cite{lin2022frozen} discards the text encoder, relying solely on temporally encoded frame features by visual~encoder. Vita-CLIP~\cite{wasim2023vita} uses shallow prompts on the text encoder, similar to~\cite{zhou2022learning}, and introduces deep temporal prompts for the visual~encoder. Recently, OAP~\cite{chatterjee2024opening} generalizes the verbs observed during training to an open vocabulary of objects with a prompt-based object encoder on egocentric videos. Our method builds on existing work while introducing an adapter architecture specifically tailored to egocentric domain, resulting in superior performance.

%% file: sec/3_method.tex
\section{X-MIC Adaptation Approach}
\label{sec:method}

\begin{figure*}[!t]
\begin{center}
\includegraphics[width=0.9\linewidth]{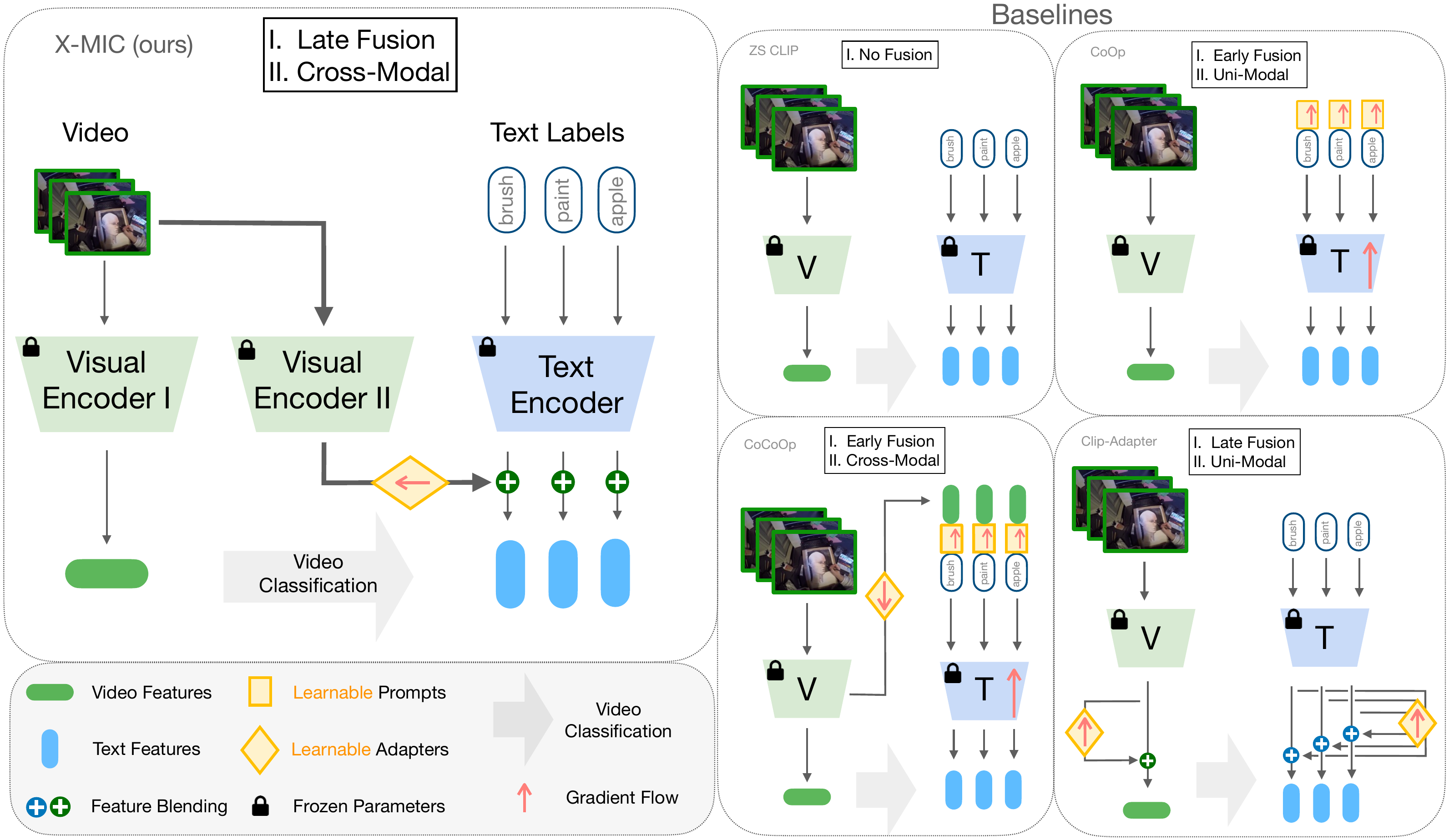}
\end{center}
\vspace{-1.4mm} 
\caption{ 
\textbf{Overview of our X-MIC method and previous adaptation methods of VLMs.} 
\underline{Baselines}: 
\textbf{No Fusion} is a standard zero-shot video classification method. The average of the frame representations is compared to text representations in the shared VL embedding space. 
\textbf{Early Fusion \& Uni-Modal} is a prompt learning method, where the learnable parameters are concatenated to text tokens and optimized through the text encoder. Subsequently, the text encoder is adapted to the new domain. 
\textbf{Early Fusion \& Cross-Modal} is an extension of Early Fusion \& Uni-Modal method, where additional learnable parameters are introduced in the form of an adapter. This adapter maps video representations to embedding space of text tokens, which are then concatenated to learnable prompts and text tokens. Memory consumption, required for forward-backwards pass through the text encoder, expands with respect to all combinations of all text-labels and videos in the batch. 
\textbf{Late Fusion \& Uni-Modal} is a method, where adaptation of both encoders is based on the feature blending of original text and video representations with the adapted corresponding representations. 
\underline{Ours:} \textbf{X-MIC} adaptation method falls in \textit{Late Fusion \& Cross-Modal} category. Adapted video features are blended with the original text features. Simple adaptation of text modality to each individual video is efficient as it does not require gradient propagation through text or video encoders. Additionally, we propose to employ Visual Encoder II, offering flexibility in utilizing various types of visual features for conditioning. Note that Visual Encoder I and II can be represented by a single visual encoder, such as the CLIP visual encoder. 
} 
\vspace{-4mm}
\label{fig:method} 
\end{figure*}

We begin by introducing the preliminaries such as classification with VLMs like CLIP and different types of VL adaptations in Sec.~\ref{sec:dual-encoder}. Then, in Sec.~\ref{sec:method_overview}, we give an overview of our adapter method for text conditioning and present our egocentric-spatio-temporal attention module.
 
\subsection{Preliminaries and Baselines on VL Adaptation}
\label{sec:dual-encoder}
Vision-language models (VLMs), such as CLIP, demonstrate effective zero-shot generalization across various downstream tasks for image recognition and third-person video recognition. However, certain domains, like egocentric videos, still face challenges due to a significant gap between web-collected and egocentric data. 

\noindent Below, we provide an overview existing prompt learning and adapter-based methods.

\myparagraph{Video Classification with VL Dual Encoders.} 
Trained on hundreds of millions of text and visual data pairs, VL dual encoders bring the two modalities together in a shared embedding space. When evaluating models pre-trained on extensive web data, a crucial metric is their ability to transfer to other downstream tasks without additional fine-tuning, a process commonly known as zero-shot evaluation. To perform zero-shot classification, one needs to propagate a set of $C$ predefined classes in the form of text, denoted as $t=$ ``Image of a $<$class$>$'' through a pre-trained text encoder $T(\cdot)$. This process extracts individual text embeddings, represented as $e_t = T(t) \in \mathcal{R}^{1 \times D}$ for each class. Subsequently, these vectors undergo $l2$ normalization, resulting in $\bar{e}_{t} = \frac{e_t}{||e_t||}$ (hereafter, the overline symbol $\bar{e}$ indicates $l2$ normalization of vector $e$). Then a matrix $\bar{E}_{T} \in \mathcal{R}^{C \times D}$ is constructed, representing a simple linear classifier, and is thus referred to as the text-based classifier. To classify an input video $v$, we sample $N$ frames, denoting the sampled frames as $v'=\{z_i\}^N$, where $z_i$ represents a frame from the video $v$. Subsequently, all sampled frames are mapped to the shared VL embedding space, using the frozen visual encoder $V(\cdot)$. Applying average pooling over the embeddings of the frames yields a single-vector video representation: $\bar{e}_{v'} = \textit{avg\_pool}(\{\overline{V(z_i)}\}^N) \in \mathcal{R}^{1 \times D}$. The video vector $\bar{e}_{v'}$ is then classified using the text-based classifier $\bar{E}_{T}$.
 
\myparagraph{No Fusion.}
We refer to frozen dual encoders $T(\cdot)$ and $V(\cdot)$ without additional adaptation as ``No Fusion'' baseline. 

\myparagraph{Early Fusion and Uni-Modal Adaptation.} 
A prompt learning-based method, CoOp~\cite{zhou2022learning}, introduces $P$ learnable vectors appended to all $C$ input text classes in the token embeddings of the textual encoder (see Fig.~\ref{fig:method}). To optimize these prompts, gradients are propagated through the frozen text encoder for $C \times P \times D$ adaptable parameters, where $D$ is the dimensionality of the tokens. This optimization remains independent of the batch size of the visual input.

\myparagraph{Early Fusion and Cross-Modal Adaptation.}
A follow-up work, CoCoOp~\cite{zhou2022conditional}, extends learnable text prompts to cross-modal prompts by introducing an adapter module of the frozen visual encoder to the token embedding space (see Fig.~\ref{fig:method}). In this architecture, each of the $C$ class-tokens are appended not only with $P$ learnable text prompts but also with individual input-conditioned prompts generated by the adapter. Optimizing these prompts for a batch of size $B$ involves propagating $B\times P \times D \times C$ gradients, making training inefficient and slow as shown in~\cite{zhou2022conditional}. 
 
\myparagraph{Late Fusion and Uni-Modal Adaptation.}
CLIP-Adapter~\cite{gao2023clip} adopts a late fusion approach as an alternative to early fusion adaptation. The text and visual encoders are followed by uni-modal adapter modules that generate adapted uni-modal feature vectors. These adapted features are then fused with the corresponding original features in the VL embedding space, subsequently optimized with the standard classification loss. This optimization is efficient due to the lightweight nature of adapters, eliminating the need for heavy text-encoder gradient propagation. 

\subsection{\our Adaptation}
\label{sec:method_overview} 
 
\myparagraph{Overview.}
We aim at achieving generalization in egocentric action recognition across domains and to novel action classes. Our \our-adaptation framework is designed to improve the alignment between frozen text representations and the egocentric visual domain directly within the VL embedding space. To adapt the text modality to the egocentric domain, we introduce a simple cross-modal text conditioning operation based on the input videos. Specifically, each \our-vector serves as an adapted video representation. We align any frozen text representation to each individual input video by a simple addition operation with the \our-vector. Consequently, text representations are adapted to individual input videos, and these adapted text embeddings are further utilized for the classification of corresponding videos into fine-grained noun and verb classes. Moreover, by introducing an egocentric-spatio-temporal attention module, we aggregate temporal information between video frames and emphasize areas around hands to enhance hand-object interactions. \our-vectors offer dual benefits: a simple and efficient cross-modal conditioning approach, and the decoupling of domain-specific knowledge from the frozen VL embedding, resulting in improved generalization on egocentric videos.

\myparagraph{\our Adaptation.} 
Our adaptation method, \our, aligns frozen text class embeddings directly to the new domain in the shared VL embedding space. During training and inference, our approach resembles zero-shot classification, as we classify frozen video representations from the original visual backbone $V(\cdot)$ using an adapted text-based classifier $\bar{E}_{T}$ tailored to each input video $v$. This enables efficient domain adaptation without the need for fine-tuning the entire model, categorizing our method as late fusion with cross-modal adaptation.
 
Specifically, for an input video $v$, we sample $N$ frames to form a sparse video sequence $v'=\{z_i\}^N$. The video sequence $v'$ is then decoded using the original visual encoder $V(\cdot)$, resulting in a single vector $\bar{e}_{v}$. Additionally, we encode the $C$ classes into the text-based classifier $\bar{E}_{T}$ as detailed in Sec.~\ref{sec:dual-encoder}.
 
To generate the \our-vector, we introduce a second frozen visual encoder, denoted as $V_{II}(\cdot)$. This secondary encoder can either be an identical copy of the original encoder $V(\cdot)$ or a distinct pre-trained encoder. In Sec.~\ref{sebsec:sota}, we demonstrate that incorporating a different type of $V_{II}(\cdot)$ can result in significant generalization improvements. For instance, DINO~\cite{oquab2023dinov2}, which is uni-modal, captures distinct characteristics~\cite{park2023self} of the visual input compared to multi-modal CLIP like models that focus solely on main objects.

We employ the second encoder $V_{II}(\cdot)$ to produce an intermediate representation of frames, denoted as $x_v = \{V_{II}(z_i)\}^N$. Before adapting the intermediate representation, we apply $l2$-normalization to the vector. See Sec.~\ref{subsec:ablations} for a detailed analysis of the impact of this normalization. Our video adapter $A(\cdot)$ incorporates a temporal aggregation module. By feeding these intermediate representations into this module, we obtain the final \our-vector for adaptation, represented as $a_{v} = A(\bar{x}_v)$.
 
Finally, to adapt the frozen text-based classifier $\bar{E}_{T}$ to the video $v$, we simply sum \our-vector with each class representation in the embedding space: $\overline{e_t + a_{v}} \in \mathcal{R}^D$, and when combined, these updated vectors form an adapted text-based classifier ${\overline{E}^{a_v}_T}$. Subsequently, we classify the video representation $\bar{e}_v$ with the adapted text-based classifier ${\overline{E}^{a_v}_T}$. The process of classification with \our-adaptation can be summarized as follows:
\begin{equation}
c = \text{argmax}_{t} <\overline{e_t + A(\overline{V_{II}(x_v)})} , \bar{e}_v >,
\label{eq:main}
\end{equation}
where $c$ represents the class with the highest similarity between the adapted text-based classifier ${\overline{E}^{a_v}_T}$ and the video $v$, and $<\cdot,\cdot>$ denotes dot product. \\

\begin{figure}[!t]
\begin{center}
\includegraphics[width=0.75\linewidth]{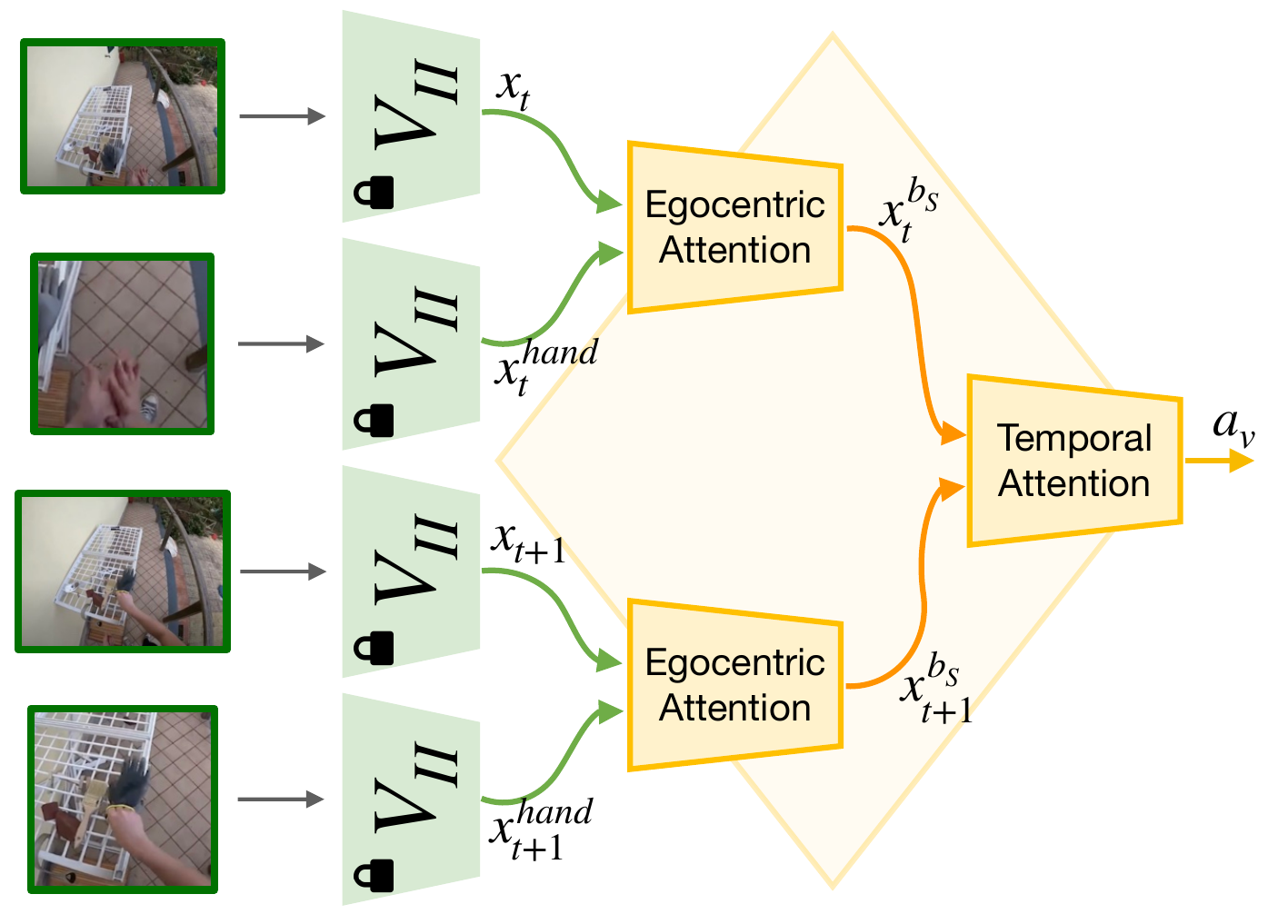}
\end{center}
\vspace{-3mm}
\caption{ 
\textbf{Ego-Spatio-Temporal Attention Module.} It takes a sequence of full frames interleaved with hand crops as input, and outputs \our vector $a_v$, representing video $v$ as a single vector for text conditioning in the shared VL embedding space. 
}
\label{fig:attention} 
\end{figure}

\myparagraph{Ego-Spatio-Temporal Attention Module.}
Our adaptation module consist of two transformer blocks $b_{S}(\cdot)$ and $b_{T}(\cdot)$ designed to aggregate different types of information. This module not only adapts each video to the shared VL embedding space but also captures egocentric video-specific spatial and temporal information, through $b_{S}(\cdot)$ and $b_{T}(\cdot)$ respectively. To capture hand-object interactions better, we introduce an attention block that focuses on regions around hands, see Fig.~\ref{fig:attention}. This involves applying self-attention between hand crops and full frames, guiding the model to emphasize information around hands, a crucial region of interest in egocentric videos. To aggregate temporal information from the video, we employ a temporal self-attention block, similar to~\cite{ju2022prompting} that updates the frame representations. Our \our-vector for adaptation is derived by applying average pooling over all updated frame representations.

Egocentric videos may include diverse backgrounds and involve significant camera motion. By focusing on the region around hands through cropping and applying self-attention to both the full frame and the cropped region, we guide our model to prioritize attention on the hands. More specifically, for each frame, we use the full frame $z_i$ and the cropped hand region $z^{hand}_i$ from the same frame. We get the intermediate representations through the video encoder, for the full frame $x_i = V_{II}(z_i)$ and the hand region $x_i^{hand} = V_{II}(z_i^{hand})$. We concatenate the two to obtain an intra-frame sequence $[x_i; x_i^{hand}] \in \mathcal{R}^{2 \times D}$. We derive the intra-frame representation $x^{b_S}_i$ by averaging the updated representations from both the full and cropped frames: 
\begin{equation}
 x^{b_S}_i = \text{avg\_pool} (b_{S}([x_i; x_i^{hand}])).
\end{equation}
 
To capture the temporal relations across frames, we apply self-attention between all frames of the video. Specifically, we use the second transformer $b_{T}(\cdot)$ block to update $x^{b_S}_i$ frame representations, which we aggregate with average pooling into our \our-vector:
\begin{equation}
 a_v = \text{avg\_pool} (b_{T}([x^{b_S}_1, x^{b_S}_2, \cdots, x^{b_S}_N])).
\end{equation} 
In this way, our adaptation module effectively incorporates egocentric video-specific spatial and temporal information into the frozen vision-language embedding space, enhancing generalization to novel classes.

%% file: sec/4_results.tex
\section{Experiments}
\label{sec:results}

\our is mainly evaluated on the cross-dataset setting between two large-scale egocentric datasets: Ego4D~\cite{Ego4D2022CVPR} and Epic-Kitchens~\cite{Damen2022RESCALING}. We also evaluate generalization performance on the small-scale EGTEA~\cite{li2018egtea} dataset. 
\input{tables/v2_verb_and_nouns}

\subsection{Datasets}
\myparagraph{Ego4D~\cite{Ego4D2022CVPR}.} We a subset of Ego4D~\cite{Ego4D2022CVPR} annotated with fine-grained noun and verb labels, specifically from the FHO benchmark which contains 521 noun and 117 verb classes. The training set consists of 64K video clips, while the testing set comprises 33K clips. The average clip duration is 8 seconds, resulting in a total of approximately 215 hours of videos, excluding irrelevant background clips. 
 
\myparagraph{Epic-Kitchens~\cite{Damen2022RESCALING}.}
We use the Epic-Kitchens100, comprising 67K video clips for training and 10K video clips for testing. The average clip length is 3.5 seconds, totaling about 70 hours, excluding irrelevant background clips. The dataset features annotations for 300 noun classes and 97 verb classes, focusing on kitchen-related topics.
 
\myparagraph{EGTEA~\cite{li2018egtea}} We use this dataset solely for model testing, given its training set of 8,000 video clips with clip length of 3.2 seconds. During inference, we combine three test splits resulting in 6K video clips in total. The dataset is annotated with fine-grained 20 verb and 54 noun classes. 
 
\input{tables/detailed_generalized_performance_mini} 
\label{sebsec:sota}

\subsection{Implementation Details}
We evaluate the generalization performance based on the adaptation of the pre-trained CLIP ViT-B/16 model, unless otherwise specified. For model training, we use AdamW with($\beta_1$, $\beta_2$) = (0.9, 0.999) and weight decay of 0.01 for 15 epochs with a fixed learning rate of 1e-6. The Transformer module $b_1$ contains 1 self-attention layer, whereas temporal attention module $b_2$ includes 2 self-attention layers. During training, we sample 16 random frames, and during evaluation we sample frames uniformly. For detecting the hand regions, we use the 100DOH~\cite{shan2020understanding} detector, extracting bounding boxes for each frame. Further implementation details are provided in the supplementary.

\myparagraph{Cross-Datasets Evaluation.}
In our work, we investigate the generalization performance of fine-grained noun and verb recognition across egocentric datasets. Our objective is two-fold: achieving strong performance within the dataset on the corresponding test set and demonstrating robust generalization to a shared dataset~(Table~\ref{tab:sota}). We compute the harmonic mean between these two to gauge the balance of different types of generalization. For instance, we train our models on Ego4D and subsequently evaluate on both the Ego4D and Epic-Kitchens test sets. We further analyse zero-shot generalization by identifying disjoint subsets of shared and novel classes across datasets (Table~\ref{tab:shared_novel_mini}). See supplementary for corresponding classes. 
 
\input{tables/egtea}

\subsection{\our Comparison to SOTA}
In Tables~\ref{tab:sota} and~\ref{tab:shared_novel_mini}, we start with comparing our method to CLIP. Notably, on both the Ego4D and Epic-Kitchens datasets, CLIP yields surprisingly low results for both verbs and nouns, in contrast to its strong performance on third-person datasets~\cite{kay2017kinetics,soomro2012ucf101}. Next, we compare our method to other adaptation methods of VLMs which have shown improvements on image and video recognition benchmarks. 

\myparagraph{Image-based Adaptation Methods.} 
First, we present a comparison to image-based adaptation models, including CoOp~\cite{zhou2022learning}, CoCoOp~\cite{zhou2022conditional}, and CLIP-Adapter~\cite{gao2023clip} which do not contain a temporal component. Our analysis in Table~\ref{tab:sota} shows that early fusion-based models like CoOp and CoCoOp exhibit limited learning capacity, resulting in poorer performance compared to other models for both nouns and verbs when evaluated within-dataset, especially on Epic-Kitchens. This aligns with earlier findings shown in \cite{chen2022adaptformer}. A late fusion-based framework, CLIP-Adapter, improves the within-dataset scores but demonstrates weaker generalization on nouns for cross dataset evaluation. However, Table~\ref{tab:shared_novel_mini} reveals that CoOp~\cite{zhou2022learning} demonstrates robustness when novel nouns and verbs are encountered even in the absence of any temporal attention module. We hypothesize that other models may be more prone to overfitting on the shared classes due to a larger number of parameters. 

\myparagraph{Video-based Adaptation Methods.} In Table~\ref{tab:sota}, we evaluate the performance of recent third-person video adaptation models, specifically A5~\cite{ju2022prompting} and Vita-CLIP~\cite{wasim2023vita}, in an egocentric scenario. Additionally, we enhance the CLIP-Adapter model by incorporating temporal attention and evaluate its effectiveness as a video model. We notice that the inclusion or exclusion of a temporal component, beyond simple averaging, has a relatively minor impact on noun recognition using CLIP-Adapter. To illustrate, when trained on the Epic-Kitchens dataset, CLIP-Adapter, with (denoted as CLIP-Adapter*) and without a temporal attention module, exhibits comparable performance in noun recognition within the dataset (EK), with scores of 33.21\% and 34.40\%, respectively. However, the role of temporal attention becomes crucial in enhancing verb recognition performance, as evidenced by consistent improvements across both datasets and all models. A5~\cite{ju2022prompting}, which combines both early fusion and temporal attention, shows poor cross-dataset generalization on nouns for both datasets, aligning with the findings reported by its authors~\cite{ju2022prompting} in the context of cross-dataset third-person video generalization. The recent SOTA model on third-person video generalization, Vita-CLIP~\cite{wasim2023vita}, demonstrates enhanced noun recognition on both datasets but exhibits lower verb recognition on Ego4D. In contrast to other video adaptation models, we decouple temporal attention from the frozen backbone and introduce \our-vector, encapsulating all temporal information. Moreover, employing cross-modal adaptation, we introduce video-specific classifiers. For each video, we create an individual text-based classifier, which is adapted with our \our-vector. Our approach demonstrates state-of-the-art generalization performance while maintaining high performance on within-dataset evaluation. Moreover, by leveraging DINO pre-trained model~\cite{oquab2023dinov2} as visual encoder $V_{II}$, we observe significant improvements on within-dataset evaluation. In Table~\ref{tab:egtea}, we present our evaluation on EGTEA. Overall, we note consistent trends across all methods. 

In Table~\ref{tab:shared_novel_mini}, we observe that video-based models perform poorly, likely due to overfitting on shared classes. Models like A5~\cite{ju2022prompting} and Vita-CLIP~\cite{wasim2023vita}, with a larger number of parameters, may be more susceptible to this issue. In contrast, our \our framework decouples the adapter module from the frozen VL embedding space, enabling enhanced generalization. Furthermore, we observe that models struggle more with generalizing on verbs than nouns, likely due to the object-centric pre-training data of the backbone models, \eg CLIP is pre-trained solely on image-text pairs.

\subsection{Ablations}
\label{subsec:ablations}
In this section, we evaluate the effectiveness of our design choices. For all ablations, we train models on Ego4D and evaluate on Ego4D and Epic-Kitchens. As backbone, we use CLIP ViT-B/16, unless otherwise specified.

\myparagraph{Ego-Spatial-Temporal Attention.} In Table~\ref{tab:ego-spatial-attn}, we demonstrate the impact of utilizing full frames, that usually includes scene context, and hand crops on the performance of egocentric videos. 
We observe that concentrating solely on hand regions enhances verb generalization, whereas the utilization of full images proves marginally more advantageous for noun generalization. When employing our proposed ego-spatial-temporal attention mechanism, we achieve a notable improvement in the harmonic mean. Specifically, there is a 1.45\% increase for nouns and a 2.14\% boost for verbs compared to using full frames. By guiding the model to consider context in relation to hand areas, our attention approach not only enhances performance within the dataset but also showcases improved cross-dataset performance. 
\input{tables/different_backbones}

\input{tables/prompts}

\myparagraph{Larger backbone.}
In Table~\ref{tab:backbones}, we assess the effectiveness of our method using a bigger CLIP model, specifically comparing the performance of CLIP ViT-L/14 with ViT-L/16. While we do not observe performance gains for within the dataset evaluations, a compelling trend emerges in cross-dataset generalization, particularly on nouns. Notably, employing the larger model ViT-L/14 results in a significant improvement of over 7\% in noun and 2.45\% in verb generalization on Epic. This encouraging outcome underscores the potential of vision transformers and suggests that further exploration and refinement of these models could yield even more substantial gains in cross-dataset generalization.

\myparagraph{Egocentric VL backbone.}
Table \ref{tab:backbones} presents an evaluation on \our-model performance using backbones CLIP and Lavila~\cite{zhao2023lavila}, which is pre-trained on text-video pairs from the Ego4D dataset in a contrastive manner. Note that the Lavila backbone initializes its model from CLIP pre-trained models. We first compare the zero-shot results from the original CLIP backbone and Lavila. Lavila demonstrates a significant improvement in noun recognition by 16.59\% on the Ego4D dataset and noun generalization to Epic by 17.18\%. While Lavila shows a decrease in verb recognition accuracy within the dataset by 2.38\% , its generalization to Epic verbs increases by 6.04\%. This outcome is surprising, as we initially expected Lavila to generalize better on verbs due to its training on an egocentric dataset, indicating a strong bias toward object-oriented pre-training strategies. We observe similar trends when our model utilizes CLIP versus Lavila as a backbone, where noun generalization increases significantly, while verb generalization slightly decreases.
 
\myparagraph{Prompts for text encoder.}
In Table \ref{tab:prompts}, we evaluate the performance of zero-shot CLIP and our model by prompting the frozen text model for classification with additional context. Our experiments include specific details like "Video of a" or indications of hands and an egocentric view. We find that zero-shot noun performance is the best with the standard ``Image of a '' context for Ego4D and without context for Epic-Kitchens. However, zero-shot verb recognition benefits from an additional context, achieving 3.29\% and 9.87\% on Ego4d and Epic-Kitchens, respectively. With our X-MIC adaptation, we observe that noun recognition remains robust to these changes, while verb recognition is sensitive and performs best when no additional context is provided highlighting the complexity of incorporating contextual information in egocentric scenarios.

\myparagraph{Importance of normalization.}
We investigate the significance of feature normalization in the embedding space in Table~\ref{tab:norm}. %
\texttt{n1} represents our default choice, involving the normalization of visual features after the $V_{II}$ encoder and before the adapter. \texttt{n2} indicates the normalization of \our-vector, i.e., visual features after our video-adapter module, prior to summation with frozen text features. Lastly, \texttt{n3} denotes the normalization of frozen text features before summation with \our-vector. The 'none' corresponds to no normalization. [\texttt{n1}] demonstrates the optimal balance between regularization and no regularization. Configurations [\texttt{n1}], [\texttt{n2,n3}], [\texttt{n1,n2,n3}], and \texttt{'none'} all yield symmetric feature magnitudes before the summation of frozen text features and \our-vector and marginally change the harmonic mean. Conversely, variations such as [\texttt{n1,n2}] and [\texttt{n1,n3}] result in imbalances during the summation of different modalities, leading to suboptimal performance.

%% file: tables/v2_verb_and_nouns.tex
\begin{table*}
\centering
\small
\begin{tabular}{l|c|ccc|ccc||ccc|ccc}
\hline
\multicolumn{2}{l|}{ } & \multicolumn{6}{c||}{Trained on Ego4D (E4D)} & \multicolumn{6}{c}{Trained on Epic-Kitchens (EK)} \\
\hline
 & ta & \multicolumn{3}{c|}{Nouns} & \multicolumn{3}{c||}{Verbs} & \multicolumn{3}{c|}{Nouns} & \multicolumn{3}{c}{Verbs} \\
\hline
\multicolumn{1}{l}{Evaluation dataset} & \multicolumn{1}{l|}{} & E4D & EK & hm & E4D & EK & hm & \multicolumn{1}{c}{EK} & E4D & hm & EK & E4D & hm \\ 
\hline
ZS CLIP & - & 5.89 & 8.74 & 7.03 & 2.18 & 4.25 & 2.88 & 8.74 & 5.89 & 7.03 & 4.25 & 2.18 & 2.88 \\
\hline 
CoOp & - & 28.22 & 10.87 & 15.70 & 22.57 & 20.42 & 21.44 & 21.56 & 9.37 & 13.06 & 30.91 & 13.35 & 18.64 \\
Co-CoOp & - & 30.00 & 9.51 & 14.44 & 21.31 & 12.99 & 16.14 & 24.23 & 9.27 & 13.41 & 34.16 & 14.17 & 20.03 \\
CLIP-Adapter & - & 30.00 & 8.95 & 13.78 & 22.82 & 19.94 & 21.28 & 33.21 & 5.73 & 9.77 & 36.70 & 16.09 & 22.37 \\
CLIP-Adapter* & \checkmark & 31.26 & 10.00 & 15.16 & 27.32 & 22.28 & 24.54 & 34.40 & 4.67 & 8.22 & 48.69 & 15.52 & 23.54 \\
A5 & \checkmark & 31.39 & 7.84 & 12.55 & 26.31 & 22.77 & 24.41 & 32.04 & 3.31 & 5.99 & 46.05 & 17.93 & \underline{25.81} \\
Vita-CLIP & \checkmark & 33.52 & 10.61 & 16.11 & 22.66 & 25.81 & 24.13 & 34.41 & 9.52 & 14.91 & 48.78 & 13.47 & 21.11 \\
\rowcolor[rgb]{0.859,0.859,0.859} X-MIC~ & \checkmark & 33.54 & 15.35 & \underline{21.06} & 28.93 & 26.48 & \underline{27.65} & 30.64 & 12.32 & \underline{17.57 } & 50.01 & 18.10 & \textbf{26.58} \\
\rowcolor[rgb]{0.859,0.859,0.859} X-MIC+DINO~ & \checkmark & 35.85 & 18.96 & \textbf{24.80 } & 28.27 & 29.49 & \textbf{28.86} & 44.07 & 11.45 & \textbf{18.17} & 53.02 & 16.01 & {24.60} \\
\hline
\end{tabular}
\caption{
\textbf{SOTA comparison on within- and cross-dataset evaluation on Ego4d and Epic Kitchens datasets. } Left: Models trained on Ego4D. Right: Models trained no Epic-Kitchens. 
Evaluation is on noun and verb classes. ta denotes temporal attention in the corresponding method, other methods apply simple average of the frames. hm stands for harmonic mean evaluation. X-MIC+DINO denotes our model with DINO~\cite{oquab2023dinov2} as Visual Encoder II. 
}
\label{tab:sota}
\end{table*}

%% file: tables/detailed_generalized_performance_mini.tex
\begin{table*}
\small
\centering 
\begin{tabular}{l|l|ccc|ccc||ccc|ccc} 
\hline
\multicolumn{1}{c}{} & & \multicolumn{6}{c||}{Trained on E4D, Evaluated on EK} & \multicolumn{6}{c}{Trained on EK, Evaluated on E4D} \\ 
\hline
 & ta & \multicolumn{3}{c|}{Nouns} & \multicolumn{3}{c||}{Verbs} & \multicolumn{3}{c|}{Nouns} & \multicolumn{3}{c}{Verbs} \\ 
\hline
Eval. subset & & shared & novel & hm & shared & novel & hm & shared & novel & hm & shared & novel & hm \\ 
\hline
ZS CLIP & - & 10.38 & 13.58 & 11.77 & 12.32 & 4.32 & 6.40 & 11.38 & 10.49 & 10.92 & 2.73 & 9.84 & 4.27 \\ 
\hline
CoOp & - & 16.86 & 16.02 & 16.43 & 25.03 & 5.97 & 9.64 & 15.77 & 10.11 & 12.32 & 20.22 & 5.69 & 8.89 \\
CoCoOp & - & 16.35 & 11.51 & 13.51 & 24.34 & 0.00 & 0.00 & 17.31 & 11.46 & \underline{13.79} & 15.32 & 6.46 & 9.09 \\
CLIP-Adapter & - & 12.46 & 5.99 & 8.09 & 21.48 & 3.09 & 5.40 & 8.72 & 7.42 & 8.02 & 19.60 & 3.00 & 5.20 \\
CLIP-Adapter* & \checkmark & 16.24 & 12.22 & 13.95 & 25.29 & 1.23 & 2.35 & 14.67 & 7.68 & 10.08 & 24.17 & 4.50 & 7.59 \\
A5 & \checkmark & 15.25 & 5.24 & 7.80 & 27.90 & 3.09 & 5.56 & 13.54 & 5.71 & 8.03 & 24.29 & 0.55 & 1.07 \\
Vita-CLIP & \checkmark & 15.84 & 6.15 & 8.86 & 27.22 & 4.11 & 7.14 & 14.60 & 9.76 & 11.69 & 16.46 & 6.58 & 9.40 \\
\rowcolor[rgb]{0.859,0.859,0.859} X-MIC & \checkmark & 20.04 & 21.51 & \underline{20.75} & 29.01 & 7.00 & \textbf{11.27} & 19.66 & 12.24 & \textbf{15.09} & 23.00 & 7.16 & \textbf{10.92} \\
\rowcolor[rgb]{0.859,0.859,0.859} X-MIC+DINO & \checkmark & 25.56 & 20.52 & \textbf{22.76} & 31.92 & 6.38 & \underline{10.63} & 18.91 & 10.67 & 13.65 & 20.55 & 6.48 & \underline{9.85} \\
\hline
\end{tabular} 
\caption{
\textbf{Zero-shot action generalization}. Left: The models are trained on Ego4D (E4D) and subsequently evaluated on Epic-Kitchnes (EK) using disjoint subsets of classes (shared and novel). Right: The models are trained on Epic-Kitchens (EK) and then evaluated in a cross-dataset manner on subsets of classes within Ego4D.
}
\label{tab:shared_novel_mini}
\end{table*}

%% file: tables/egtea.tex
\begin{table*}[t]
\parbox{.5\linewidth}{ 
\scalebox{1}{ 
\centering
\setlength{\tabcolsep}{2mm}{
\resizebox{1.05\columnwidth}{!}{
\centering
\begin{tabular}{l|ccc|ccc}
\hline
 & \multicolumn{3}{c|}{Nouns } & \multicolumn{3}{c}{Verbs } \\
\hline
\multicolumn{1}{c|}{} & E4D & EGTEA & hm & E4D~ & EGTEA & hm \\
\hline 
ZS CLIP & 5.89 & 19.70 & 9.07 & 2.18 & 18.71 & 3.51 \\ 
CoOp & 29.23 & 23.90 & 26.29 & 22.57 & 26.45 & 24.35 \\
Co-CoOp & 29.85 & 27.90 & 28.84 & 21.31 & 27.74 & 24.10 \\
CLIP-Adapter & 30.00 & 21.41 & 24.98 & 22.82 & 26.51 & 24.52 \\
CLIP-Adapter * & 29.18 & 22.40 & 25.34 & 27.32 & 26.57 & 26.93 \\
A5 & 33.50 & 23.70 & 27.76 & 26.31 & 28.03 & 27.14 \\
Vita-CLIP & 33.52 & 17.24 & 22.76 & 22.66 & 27.63 & 24.89 \\
\rowcolor[rgb]{0.859,0.859,0.859} X-MIC & 33.54 & 29.21 & \textbf{31.21} & 28.93 & 31.41 & \textbf{30.12} \\
\hline
\end{tabular}
}}}
\caption{
\textbf{SOTA comparison on EGTEA.} The model is trained on Ego4D dataset and evaluated in a zero-shot manner on EGTEA. 
}
\label{tab:egtea}
}
\hfill
\parbox{.45\linewidth}{
\centering
\scalebox{1}{ 
\centering
\setlength{\tabcolsep}{2.7mm}{
\resizebox{0.93\columnwidth}{!}{
\begin{tabular}{c|ccc||ccc} 
\hline
 & \multicolumn{3}{c||}{Nouns} & \multicolumn{3}{c}{Verbs} \\
 \hline
 & \multicolumn{1}{c}{E4D} & \multicolumn{1}{c}{EK} & \multirow{1}{*}{hm} & \multicolumn{1}{c}{E4D} & \multicolumn{1}{c}{EK} & \multirow{1}{*}{hm} \\ 
\hline
F & 31.68 & 14.20 & 19.61 & 27.19 & 24.02 & 25.51 \\
H & 31.35 & 14.02 & 19.37 & 26.32 & 26.59 & 26.46 \\ 
F+H & 33.54 & 15.35 & \textbf{21.06} & 28.93 & 26.48 & \textbf{27.65} \\
\hline
\end{tabular}
}}}
\caption{
\textbf{Influence of Ego-Spatial-Temporal attention.} 
F denotes full frames, H denotes hand crops. F+H correspond to our proposed attention module. All models share the same architecture of the temporal attention module.
}
\label{tab:ego-spatial-attn}
}
\vspace{-2mm}
\end{table*}

%% file: tables/different_backbones.tex
\begin{table*}[]
\parbox{.68\linewidth}{
\small 
\centering
\begin{tabular}{c|c|c|ccc|ccc} 
\hline
\multicolumn{1}{c}{} & \multicolumn{1}{c}{} & & \multicolumn{3}{c|}{Nouns } & \multicolumn{3}{c}{Verbs } \\ 
\hline
\multicolumn{3}{c|}{Evaluation dataset} & E4D & EK & hm & E4D & EK & hm \\ 
\hline
\multirow{4}{*}{CLIP} & \multirow{2}{*}{\footnotesize{ViT-B/16}} & ZS & 5.89 & 8.74 & 7.03 & 2.18 & 4.25 & 2.88 \\
 & & X-MIC & 33.54 & 15.35 & 21.06 & 28.93 & 26.48 & 27.65 \\ 
\cline{2-9}
& \multirow{2}{*}{\footnotesize{ViT-L/14}} & ZS & 8.40 & 13.88 & 10.46 & 8.57 & 9.70 & 9.10 \\
 & & X-MIC & 33.75 & 22.46 & 26.97 & 28.13 & 28.93 & 28.52 \\ 
 
\hline 
\multirow{2}{*}{Lavila} & \multirow{2}{*}{\footnotesize{ViT-L/14}} & ZS & 24.99 & 31.06 & 27.69 & 6.19 & 15.74 & 8.88 \\
 & & X-MIC & 35.18 & 34.97 & 35.08 & 12.28 & 24.66 & 16.37 \\
\hline
\end{tabular}
\caption{\textbf{Influence of different backbones.} We compare the performance of CLIP ViT-L/14 with ViT-B/16. Additionally, we provide a comparison of CLIP backbone, pre-trained on text-image pairs, to Lavila backbone, pre-trained on pairs of egocentric videos and narrations from full Ego4D. ZS denotes zero-shot evaluation. X-MIC denotes the evaluation of our method with the corresponding backbones (CLIP or Lavila) \textit{without} additional DINO backbone.} 
\label{tab:backbones}
}
\hfill
\parbox{.3\linewidth}{
\centering
\scalebox{1}{ 
\centering
\setlength{\tabcolsep}{2mm}{
\resizebox{0.6\columnwidth}{!}{
\begin{tabular}{c|ccc} 
\hline
norm & \multicolumn{3}{c}{Nouns} \\ 
\hline
 & \multicolumn{1}{c}{E4D} & \multicolumn{1}{c}{EK} & hm \\ 
\hline
n1 & 33.54 & 15.35 & \textbf{21.06} \\
\hline
none & 32.64 & 14.34 & 19.92 \\
n2,n3 & 32.74 & 14.59 & 20.19 \\
n1,n2,n3 & 31.99 & 14.49 & 19.95 \\
\hline
n1,n2 & 15.81 & 12.3 & 13.83 \\
n1,n3 & 12.12 & 11.34 & 11.71 \\
\hline
\end{tabular}
}}}
\caption{\textbf{Influence of feature normalization}. [n1] stands for $l2$-norm of features after $V_{II}$ encoder and before the adapter (our default). [n2] denotes $l2$-norm of \our vector before sum, [n3] denotes $l2$-norm of text features before sum. 
}
\label{tab:norm} 
}
\end{table*}

%% file: tables/prompts.tex
\begin{table*}
\centering
\scalebox{1}{ 
\centering
\setlength{\tabcolsep}{2.1mm}{
\resizebox{0.97\linewidth}{!}{
\begin{tabular}{l|cc|ccc|cc|ccc} 
\hline
\multicolumn{1}{c|}{} & \multicolumn{5}{c|}{Nouns} & \multicolumn{5}{c}{Verbs} \\ 
\cline{2-11}
 & \multicolumn{2}{c|}{ZS} & \multicolumn{3}{c|}{X-MIC} & \multicolumn{2}{c|}{ZS} & \multicolumn{3}{c}{X-MIC} \\ 
\hline
\multicolumn{1}{c|}{prompts} & E4D & EK & E4D & EK & hm & E4D & EK & E4D & EK & hm \\ 
\hline
$<$class$>$ & 5.89 & \textbf{8.74} & 33.54 & 15.35 & \underline{21.06} & 2.18 & 4.25 & 28.93 & 26.48 & \textbf{27.65} \\
Image of a $<$class$>$ & \textbf{10.52} & 6.75 & 32.31 & 14.81 & 20.31 & \underline{3.28} & 5.40 & 28.56 & 25.98 & \underline{27.21} \\
Video of a $<$class$>$ & \underline{10.32} & 6.80 & 32.62 & 14.77 & 20.33 & 2.93 & 5.97 & 28.14 & 22.70 & 25.13 \\
Egocentric image a $<$class$>$ & 9.61 & 7.11 & 32.09 & 15.65 & 21.04 & 2.98 & 3.83 & 28.58 & 24.02 & 26.10 \\
Image of a hand holding a $<$class$>$ & 10.09 & 6.32 & 32.92 & 14.28 & 19.92 & \textbf{3.29} & \textbf{9.87} & 27.53 & 19.33 & 22.71 \\
Egocentric image of a hand holding $<$class$>$ & 9.23 & \underline{6.86} & 33.29 & 15.83 & \textbf{21.45} & 2.41 & \underline{6.24} & 27.66 & 16.94 & 21.01 \\
\hline
\end{tabular}
}}}
\caption{\textbf{
Influence of prompting the frozen text model with additional context.} ZS denotes zero-shot CLIP evaluation. 
Noun recognition is robust to contextual variations, while verb recognition performs best without additional context.
}
\label{tab:prompts}
\vspace{-3mm}
\end{table*}

%% file: sec/5_conclusion.tex
\section{Conclusions \& Limitations}
\label{sec:conclusion}
We have introduced X-MIC, a simple yet effective cross-modal adaptation framework for VLMs, that injects egocentric video information into the frozen VL embedding, achieving significant improvements in fine-grained cross-dataset egocentric recognition of nouns and verbs. Moreover, X-MIC vectors offer decoupling of the domain-specific knowledge from the frozen VL embedding. This allows to explore different visual backbones for text conditioning directly in the embedding space, showing improved generalization. It is important to note that our method focuses solely on video classification and does not encompass text-vision tasks like text-to-video retrieval, which would necessitate using text-conditioned videos instead of our video-conditioned text representations. We plan to explore this direction in future work.

%% file: sec/X_suppl.tex
\setcounter{page}{1}
\maketitlesupplementary
 
In this supplementary, we provide additional implementation details in Sec.~\ref{sec:implemetation_details}, extend the discussion on generalization performance to shared and novel action classes in Sec.~\ref{sec:generalization}, explore the synergies between various adaptation techniques and \our framework in Sec.~\ref{sec:complementarity}, and present additional ablation experiments of our \our framework in Sec.~\ref{sec:ablations}. 
 
\input{tables/scale} 
\input{tables/different_backbones_supmat}
\input{tables/normalization_supmat}
\input{tables/fusions_supmat}
\input{tables/N_frames_supmat}

\input{tables/attention_supmat}
\section{Implementation Details}
\label{sec:implemetation_details}
\myparagraph{Transformer Block.} In our Ego-spatio-temporal attention module, we utilize a sequence of transformer blocks $b_S$ and $b_T$ to capture spatial and temporal dependencies, respectively. The general structure of the transformer block given input tensors $x$ of dimensionality $D$ is depicted in Alg.~\ref{alg:transformer}. LN denotes LayerNorm~\cite{ba2016layer}, MHA denotes multi-head attention with 8 heads~\cite{vaswani2017attention}, and MLP denotes 2-layer multi-layer perception with the bottleneck $D / 4$ and QuickGELU~\cite{hendrycks2016gaussian} as an activation function. 

\begin{algorithm}
\caption{Transformer Block}\label{alg:transformer}
\begin{algorithmic}
\Require $x \in \mathcal{R}^D$ 
\State $x \gets x + MHA(LN(x))$
\State $x \gets x + MLP(LN(x))$
\end{algorithmic}
\end{algorithm}
\noindent We note that $b_S$ consist of one transformer block and $b_T$ includes two sequential transformer blocks.

\myparagraph{Data.} 
For augmentations, we exclusively employ frame flipping during training. Furthermore, for the CLIP backbone, we resize frames so that the shortest side is 224, followed by a center crop of 224x224 and normalization. For the Lavila backbone, frames are directly resized to 224x224. 
 
For Epic-Kitchens, we use  the provided annotation boundaries to define action clips. Conversely, in the Ego4D FHO challenge, the boundaries for each clip are initially set as 8 seconds. However, upon observation, we note that actions typically span a duration shorter than 8 seconds, prompting us to uniformly shorten all clips to 4 seconds.

\section{Zero-Shot Generalization Discussion}
\label{sec:generalization}
In Table~\ref{tab:shared_novel_mini}, we present comprehensive cross-dataset results showcasing the generalization performance on both the Epic-Kitchens and Ego4D datasets. Notably, the Epic-Kitchens dataset is exclusive to kitchen-related scenes and actions, while the Ego4D dataset encompasses a diverse range of daily activities. 
We distinguish between subsets of shared and novel classes and provide additional details in the following paragraph.

\myparagraph{Shared-Novel Classes.}
We categorize classes as "shared" when there is an exact match in their names across datasets. For noun classes in Ego4D and Epic Kitchens, there exist 163 such shared classes, including examples like ``apple'', ``toaster'' or ``washing machine''. Consequently, the set of "novel" noun classes for Ego4D comprises 358 classes, encompassing items such as ``transistor'', ``ambulance'' and ``stroller''. In contrast, the "novel" noun classes for Epic-Kitchens total 137 and predominantly represent more detailed kitchen-related categories such as ``mint'', ``onion ring'', ``scale''. In the domain of verb classes for both Ego4D and Epic Kitchens, we identify 51 \textit{shared} classes, including actions such as ``hold'', ``hang'' or ``attach''. This results in 66 \textit{novel} verb classes for Ego4D with examples like ``park'', ``repair'' and ``wave''. On the other hand, \textit{novel} verb classes for Epic-Kitchens amount to 46, primarily encompassing more detailed kitchen-related actions like ``slide'', ``stab'', ``unfreeze''. For a comprehensive list of classes, refer to the detailed class separation in Sec. \ref{sec:list-of-classes}.

\section{Complementary of \our}
\label{sec:complementarity} 
In Table~\ref{tab:fusions}, we illustrate the compatibility of our framework with other adaptation methods. Early fusion methods, where the uni-modal (U) method corresponds to CoOp~\cite{zhou2022learning} and the cross-modal (X) method corresponds to CoCoOp~\cite{zhou2022conditional}, demonstrate enhanced performance in both within- and cross-dataset evaluations. However, the integration of our framework with late fusion uni-modal adapters (Tt and Vv) does not further enhance the generalization while maintaining the overall high performance.
 
\section{Ablations of \our}
\label{sec:ablations}
In this section, we conduct additional ablation experiments to validate the efficacy of our design choices.

\input{tables/temporal_att}

\myparagraph{Temporal Attention.} 
In Table \ref{tab:temp_attention}, we present an ablation where we keep our ego-spatial attention but replace the temporal module with a simple average over frames. The results demonstrate significant improvements in verb recognition on Ego4D by 6.28\%, and in verb generalization to Epic by 6.29\% when employing temporal modeling. This aligns with expectations, as the temporal component encodes movements. We also observe improvements in noun recognition and generalization, indicating that recognizing nouns in an egocentric dataset requires more than just appearance; motion encoding is also beneficial \eg for recognizing "slicing an apple" action. 

\myparagraph{Number of frames.} 
In Table~\ref{tab:num_frames_supmat}, we showcase the influence of the number of frames sampled during both training and evaluation. Notably, the performance peaks with 16 and 8 sampled frames. Conversely, sampling only two frames per clip significantly diminishes performance across all classes. 

\myparagraph{Scale of \our-vector.} 
In Eq. 1 in our main paper, we extend our analysis to validate the importance of the scale of the \our-vector. To be specific, according to Eq. 1 in our main paper is the following: 
\begin{equation}
c = \text{argmax}_{t} <\overline{e_t + \alpha A(\overline{V_{II}(x_v)})} , \bar{e}_v >,
\label{eq:main2}
\end{equation}
where $\alpha$ is a scale factor. In Table \ref{tab:scale_supmat}, we vary the scale factor $\alpha$ from 0.1 to 5 and observe that higher values result in improved performance, particularly in the evaluation of verbs both within and across datasets.

\myparagraph{Ego-Spatial-Temporal Attention.} 
In Table \ref{tab:ego-spatial-attn-supmat}, we provide supplementary results to highlight the impact of our ego-spatial-temporal attention module. We note consistent performance across models trained on Epic-Kitchens and Ego4D, see Table 3 in our main paper.

\myparagraph{Different backbones. } 
Table \ref{tab:backbones_supmat} presents a comparison between CLIP ViT-B/16 and ViT-L/14 models when trained on Epic-Kitchens. Furthermore, we evaluate the performance of image-text pre-training (CLIP model) and video egocentric pre-training (Lavila). Our findings f those outlined in Table 5 in our main paper. 

\myparagraph{Importance of normalization.} 
Table \ref{tab:norm_supmat} offers supplementary results to complement those in Table 6 our main paper. 
The outcomes closely align with the findings presented our main paper. 
 
\section{List of Classes}
\label{sec:list-of-classes}
\myparagraph{Shared Nouns:}
\begin{verbatim}
spoon;	plate;	knife;	pan;	lid;	bowl;	drawer;	sponge;	glass;	hand;	fridge;	cup;	fork;	bottle;	onion;	cloth;	chopping board;	bag;	spatula;	container;	dough;	water;	meat;	pot;	potato;	oil;	cheese;	bread;	food;	tray;	pepper;	colander;	carrot;	tomato;	kettle;	pasta;	oven;	sauce;	paper;	garlic;	towel;	egg;	rice;	mushroom;	chicken;	coffee;	glove;	leaf;	sink;	milk;	jug;	salad;	dishwasher;	cucumber;	peach;	flour;	courgette;	filter;	butter;	scissors;	chopstick;	blender;	mat;	spice;	sausage;	napkin;	microwave;	pizza;	button;	stock;	grater;	ladle;	yoghurt;	cereal;	broccoli;	brush;	lemon;	juicer;	light;	squash;	leek;	fish;	lettuce;	seed;	foil;	washing machine;	corn;	soup;	clip;	lighter;	ginger;	tea;	nut;	vinegar;	rolling pin;	pie;	burger;	book;	tongs;	cream;	banana;	paste;	plug;	teapot;	floor;	lime;	bacon;	sandwich;	phone;	thermometer;	orange;	basket;	tablet;	cake;	avocado;	chair;	pancake;	toaster;	apple;	chocolate;	ice;	handle;	pea;	yeast;	coconut;	spinach;	apron;	grape;	kale;	wire;	asparagus;	mango;	kiwi;	bean;	whisk;	remote control;	label;	celery;	cabbage;	ladder;	battery;	pear;	funnel;	wall;	strawberry;	shelf;	straw;	cork;	window;	bar;	heater;	watch;	melon;	popcorn;	candle;	balloon;	computer;	key;	pillow;	pen;	plum;	tape;	camera;
\end{verbatim}

\myparagraph{Novel Nouns Ego4D:}
\begin{verbatim}
arm;	artwork;	awl;	axe;	baby;	baking soda;	ball;	ball bearing;	baseboard;	bat;	bat;	bathtub;	batter;	bead;	beaker;	bed;	belt;	bench;	berry;	beverage;	bicycle;	blanket;	block;	blower;	bolt extractor;	bookcase;	bracelet;	brake;	brake pad;	branch;	brick;	broom;	bubble gum;	bucket;	buckle;	butterfly;	cabinet;	calculator;	caliper;	can opener;	canvas;	car;	card;	cardboard;	carpet;	cart;	cat;	ceiling;	cello;	cement;	chaff;	chain;	chalk;	chip;	chip;	chip;	chisel;	cigarette;	circuit;	clamp;	clay;	clock;	coaster;	coffee machine;	comb;	cooker;	cookie;	corner;	countertop;	crab;	cracker;	crayon;	crochet;	crowbar;	curtain;	cushion;	cutter;	decoration;	derailleur;	detergent;	dice;	dog;	door;	doorbell;	dough mixer;	doughnut;	dress;	drill;	drill bit;	drum;	dumbbell;	dust;	duster;	dustpan;	eggplant;	engine;	envelope;	eraser;	facemask;	fan;	faucet;	fence;	file;	filler;	fishing rod;	flash drive;	flower;	foam;	foot;	fries;	fuel;	game controller;	garbage can;	gasket;	gate;	gauge;	gauze;	gear;	generator;	glasses;	glue;	glue gun;	golf club;	gourd;	grain;	grapefruit;	grass;	grill;	grinder;	guava;	guitar;	hair;	hammer;	hanger;	hat;	hay;	haystack;	head;	headphones;	helmet;	hinge;	hole;	horse;	hose;	house;	ice cream;	ink;	iron;	jack;	jacket;	ketchup;	keyboard;	leash;	leg;	lever;	lock;	lubricant;	magnet;	manure;	mask;	matchstick;	medicine;	metal;	microscope;	mirror;	mixer;	mold;	money;	mop;	motorcycle;	mouse;	mouthmower;	multimeter;	nail cutter;	nail gun;	nail polish;	necklace;	needle;	net;	nozzle;	nut;	okra;	paddle;	paint;	paint roller;	paintbrush;	palette;	panel;	pantspapaya;	pastry;	peanut;	pedal;	peel;	peeler;	peg;	pencil;	photo;	piano;	pickle;	picture;	pilot jet;	pin;	pipe;	planer;	plant;	playing cards;	plier;	pole;	pot;	pump;	pumpkin;	purse;	puzzle or game piece;	rack;	radio;	rail;	rake;	razor blade;	ring;	rod;	root;	rope;	router;	rubber band;	ruler;	sand;	sander;	sandpaper;	saw;	scarf;	scoopscraper;	screw;	screwdriver;	sculpture;	seasoning;	set square;	sewing machine;	sharpener;	shears;	sheet;	shell;	shirt;	shoe;	shovel;	shower head;	sickle;	sieve;	sketch pad;	skirt;	slab;	snorkel;	soap;	sock;	socket;	sofa;	soil;	solder iron;	spacer;	speaker;	sphygmomanometer;	spirit level;	spray;	spring;	squeezer;	stairs;	stamp;	stapler;	steamer;	steering wheel;	stick;	sticker;	stone;	stool;	stove;	strap;	string;	stroller;	switch;	syringe;	table;	taco;	tape measure;	television;	tent;	test tube;	tie;	tile;	timer;	toilet;	toilet paper;	toolbox;	toothbrush;	toothpick;	torch;	toy;	tractor;	trash;	treadmill;	tree;	trimmer;	trowel;	truck;	tweezer;	umbrella;	undergarment;	vacuum;	vacuum cleaner;	valve;	vase;	video game;	violin;	wallet;	wallpaper;	watermelon;	weighing scale;	welding torch;	wheat;	wheel;	wheelbarrow;	windshield;	wiper;	wood;	worm;	wrapper;	wrench;	yam;	zipper;	zucchini;	ambulance;	back;	bamboo;	bandage;	baton;	bird;	brownie;	cash register;	cassava;	cocoa;	cow;	cupcake;	drone;	earplug;	hotdog;	marble;	person;	pipette;	plunger;	printer;	putty;	racket;	ratchet;	road;	scaffold;	stereo;	transistor;
\end{verbatim}

\vspace{4mm}
\myparagraph{Novel Nouns Epic-Kitchens:}
\begin{verbatim}
tap;	cupboard;	washing liquid;	box;	hob;	package;	bin;	salt;	jar;	top;	skin;	coffee maker;	rubbish;	cutlery;	can;	heat;	aubergine;	chilli;	mixture;	clothes;	tofu;	olive;	potato peeler;	cover;	kitchen towel;	vegetable;	plastic wrap;	sugar;	biscuit;	wrap;	scale;	rest;	drying rack;	alarm;	salmon;	freezer;	spreads;	cap;	curry;	oatmeal;	spring onion;	holder;	powder;	egg shell;	pork;	oregano;	food processor;	recipe;	liquid;	pak choi;	slow cooker;	utensil;	noodle;	salami;	kitchen;	tuna;	omelette;	parsley;	salad spinner;	presser;	coriander;	bottle opener;	lentil;	blueberry;	extractor fan;	salt cellar;	hummus;	juice;	green bean;	knob;	wine;	pith;	fishcakes;	raisin;	basil;	paprika;	caper;	drink;	stalk;	turmeric;	whetstone;	thyme;	lady finger;	beef;	blackberry;	slicer;	hoover;	breadstick;	roll;	cocktail;	crisp;	beer;	dust pan;	washing powder;	backpack;	cumin;	pizza cutter;	air;	quorn;	almond;	tv;	egg scotch;	stand;	vide sous machine;	masher;	hand guard;	shrimp;	fruit;	artichoke;	cherry;	sprout;	sushi mat;	crab stick;	onion ring;	pestle;	gin;	mint;	lemon grass;	rubber;	gherkin;	breadcrumb;	cinnamon;	dumpling;	rosemary;	power;	syrup;	pineapple;	sheets;	soda;	raspberry;	airer;	turkey;	face;	whiskey;	kitchen door;	cd;	vanilla extract;
\end{verbatim}

\myparagraph{Shared Verbs:}
\begin{verbatim}
take;	put;	wash;	open;	close;	insert;	turn on;	cut;	turn off;	pour;	mix;	move;	remove;	throw;	shake;	scoop;	adjust;	squeeze;	peel;	press;	turn;	scrape;	fill;apply;	fold;	break;	pull;	lift;	hold;	unroll;	hang;	sprinkle;	spray;	roll;	search;	stretch;	knead;	divide;	sharpen;	water;	attach;	wear;	measure;	unscrew;	grate;	screw;	serve;	uncover;	lock;	carry;	mark;
\end{verbatim}

\myparagraph{Novel Verbs Ego4D:}
\begin{verbatim}
arrange;	blow;	catch;	clap;	clean;	climb;	consume;	count;	cover;	crochet;	detach;	dig;	dip;	draw;	drill;	drive;	enter;	feed;	file;	fry;	give;	grind;	hit;	inspect;	iron;	kick;	knit;	loosen;	mold;	operate;	pack;	paint;	park;	pet;	plant;	play;	point;	pump;	push;	read;	repair;	sand;	scroll;	sew;	shuffle;	sieve;	sit;	smooth;	stand;	step;	stick;	swing;	talk;	tie;	tighten;	tilt;	touch;	unfold;	untie;	walk;	weld;	wipe;	write;	zip;	watch;	wave; 
\end{verbatim}

\myparagraph{Novel Verbs Epic-Kitchens:}
\begin{verbatim}
dry;	empty;	flip;	check;	scrub;	pat;	eat;	wrap;	filter;	look;	sort;	rip;	cook;	add;	crush;	set;	feel;	rub;	soak;	brush;	drop;	drink;	slide;	gather;	turn down;	coat;	transition;	increase;	wait;	lower;	form;	smell;	use;	let go;	finish;	stab;	unwrap;	choose;	flatten;	switch;	season;	unlock;	prepare;	bake;	bend;	unfreeze;
\end{verbatim}

%% file: tables/scale.tex
\begin{table*}[]
\parbox{.5\textwidth}{
\begin{tabular}{c|ccc|ccc} 
\hline
 & \multicolumn{3}{c|}{Nouns} & \multicolumn{3}{c}{Verbs} \\ 
\hline
$\alpha$ & E4D & EK & hm & E4D & EK & hm \\ 
\hline
0.1 & 31.23 & 14.46 & 19.77 & 22.85 & 24.91 & 23.83 \\
0.5 & 32.43 & 14.40 & 19.94 & 26.81 & 23.80 & 25.21 \\ 
1.0 & 33.54 & 15.35 & \textbf{21.06} & 28.93 & 26.48 & \textbf{27.65} \\
2.0 & 33.20 & 14.73 & 20.40 & 28.14 & 26.39 & 27.24 \\
5.0 & 32.29 & 14.24 & 19.77 & 27.50 & 27.00 & 27.25 \\
\hline
\end{tabular}
\caption{\textbf{Influence of scale of \our vector.} Trained on Ego4D.}
\label{tab:scale_supmat}
}
\quad
\parbox{.5\textwidth}{
\begin{tabular}{c|ccc|ccc}
\hline
 & \multicolumn{3}{c|}{Nouns} & \multicolumn{3}{c}{Verbs} \\ \hline
 & EK & E4D & hm & EK & E4D & hm \\ \hline
F & 24.89	&	12.48	&	16.62 & 42.10	&	18.86	&	26.05	\\
H & 31.13	&	11.58	&	16.88 & 44.20	&	18.49	&	26.07	\\
F+H & 30.64 & 12.32 & \textbf{17.57} & 50.01 & 18.10 & \textbf{26.58} \\ \hline
\end{tabular}
\caption{\textbf{Influence of Ego-Spatial-Temporal attention.} 
F denotes full frames, H denotes hand crops. F+H correspond to our proposed attention module. All models share the same architecture of the temporal attention module. Trained on Epic-Kitchens.}
\label{tab:ego-spatial-attn-supmat}
}
\vspace{2mm}
\end{table*}

%% file: tables/different_backbones_supmat.tex
\begin{table*}[h!]
\small
\centering
\begin{tabular}{c|c|c|ccc|ccc}
\hline
\multicolumn{1}{c}{} & \multicolumn{1}{c}{} & & \multicolumn{3}{c|}{Nouns } & \multicolumn{3}{c}{Verbs } \\
\hline
\multicolumn{3}{c|}{Evaluation dataset} & EK & E4D & hm & EK & E4D & hm \\
\hline
\multirow{4}{*}{CLIP} & \multirow{2}{*}{ViT-L/16} & Zero-Shot & 8.74 & 5.89 & 7.03 & 4.25 & 2.18 & 2.88 \\
 & & X-MIC & 30.64 & 12.32 & 17.57 & 50.01 & 18.10 & 26.58 \\
\cline{2-9}
& \multirow{2}{*}{ViT-L/14} & Zero-Shot & 13.88 & 8.40 & 10.46 & 9.70 & 8.57 & 9.10 \\
 & & X-MIC & 39.02	&	14.24	&	20.86 & 48.12	&	18.83	&	27.07	\\

\hline
\multirow{2}{*}{Lavila} & \multirow{2}{*}{ViT-L/14} & Zero-Shot & 31.06 & 24.99 & 27.69 & 15.74 & 6.19 & 8.88 \\
 & & X-MIC & 41.78	&	29.62	&	34.67 & 46.14	&	9.35	&	15.54	\\
\hline
\end{tabular}
\caption{\textbf{Influence of different backbones.} We compare the performance of CLIP ViT-L/14 with ViT-L/16. Additionally, we provide a comparison of CLIP backbone, pretrainied on text-image pairs, to Lavila backbone, pretrained on pairs of egocentric videos and narrations from full Ego4D. Trained on Epic-Kitchens (EK). }
\label{tab:backbones_supmat}
\end{table*}

%% file: tables/normalization_supmat.tex
\begin{table*}[h!]
\centering
\begin{tabular}{c|ccc|ccc||ccc|ccc} 
\hline
\multirow{3}{*}{norm} & \multicolumn{6}{c||}{Trained on Ego4D (E4D)} & \multicolumn{6}{c}{Trained on Epic-Kitchens (EK)} \\ 
\cline{2-13}
 & \multicolumn{3}{c|}{Nouns} & \multicolumn{3}{c||}{Verbs} & \multicolumn{3}{c|}{Nouns} & \multicolumn{3}{c}{Verbs} \\ 
\cline{2-13}
 & E4D & EK & hm & E4D & EK & hm & EK & E4D & hm & EK & E4D & hm \\ 
\hline
n1 & 33.54 & 15.35 & \textbf{21.06} & 28.93 & 26.48 & \textbf{27.65} & 30.64 & 12.32 & \textbf{17.57} & 50.01 & 18.10 & 26.58 \\ 
\hline
none & 32.64 & 14.34 & 19.92 & 27.79 & 25.76 & 26.74 & 32.54 & 10.34 & 15.69 & 43.25 & 19.53 & \textbf{26.91} \\
n2,n3 & 32.74 & 14.59 & 20.19 & 27.55 & 24.49 & 25.93 & 34.08 & 10.48 & 16.03 & 49.32 & 16.49 & 24.71 \\
n1,n2,n3 & 31.99 & 14.49 & 19.95 & 24.30 & 22.69 & 23.47 & 32.46 & 10.21 & 15.54 & 49.02 & 18.05 & 26.39 \\ 
\hline
n1,n2 & 15.81 & 12.3 & 13.83 & 24.3 & 22.69 & 23.47 & 19.88 & 8.14 & 11.55 & 19.72 & 8.41 & 11.79 \\
n1,n3 & 12.12 & 11.34 & 11.71 & 22.88 & 18.49 & 20.46 & 16.16 & 8.72 & 11.33 & 23.68 & 17.68 & 20.24 \\
\hline
\end{tabular}
\caption{\textbf{Influence of feature normalization}. Extended Table 6(main). 
[\texttt{n1}] corresponds to the normalization of visual features after the $V_{II}$ encoder and before the adapter and demonstrates an optimal balance between normalization and no normalization. [\texttt{n2}] corresponds to the normalization of the X-MIC vector before summation with text representation. [\texttt{n3}] corresponds to the normalization of text representation before summation with the X-MIC vector.}
\label{tab:norm_supmat}
\end{table*}

%% file: tables/fusions_supmat.tex
\begin{table*}[h!]
\centering
\begin{tabular}{l|c|ccc|ccc}
\hline
 & U/X & \multicolumn{3}{c|}{Nouns} & \multicolumn{3}{c}{Verbs} \\
 \cline{3-8}
 & -Modal & E4D & EK & hm & E4D & EK & hm \\ \hline
Early Fusion & U & 28.22 & 10.87 & 15.70 & 22.57 & 20.42 & 21.44 \\
+X-MIC & U+X & 29.83 & 11.94 & 17.05 & 24.99 & 22.72 & 23.80 \\ \hline
Early Fusion & X & 30.00 & 9.51 & 14.44 & 21.31 & 12.99 & 16.14 \\ 
+X-MIC & X+X & 30.33 & 10.34 & 15.42 & 25.53 & 25.06 & 25.29 \\ \hline
X-MIC & X & 33.54 & 15.35 & 21.06 & 28.93 & 26.48 & 27.65 \\
+ Tt & X+U & 33.66 & 14.82 & 20.58 & 28.41 & 26.69 & 27.52 \\
+ Vv & X+U & 32.75 & 15.20 & 20.77 & 28.36 & 25.85 & 27.05 \\
+ Tt + Vv & X+U & 33.23 & 15.13 & 20.79 & 28.20 & 26.29 & 27.21 \\ \hline
\end{tabular}
\caption{\textbf{\our framework with other adaptation methods.} U denotes uni-modal methods, X denotes cross-modal methods. Tt denotes text uni-modal adapter for late fusion, Vv similarly denotes video uni-modal adapter for late fusion. Trained on Ego4D (E4D). }
\label{tab:fusions}
\end{table*}

%% file: tables/N_frames_supmat.tex
\begin{table}
\centering
\scalebox{1}{
\centering
\setlength{\tabcolsep}{1.4mm}{
\resizebox{0.99\columnwidth}{!}{
\begin{tabular}{c|ccc|ccc}
\hline
\multirow{2}{*}{\# frames} & \multicolumn{3}{c|}{Nouns} & \multicolumn{3}{c}{Verbs} \\
\cline{2-7}
 & E4D & EK & hm & E4D & EK & hm \\
\hline
32 & 33.02 & 14.77 & 20.41 & 28.56 & 25.34 & 26.85 \\
16 & 33.54 & 15.35 & \textbf{21.06} & 28.93 & 26.48 & \textbf{27.65} \\
8 & 32.06 & 14.82 & 20.27 & 27.30 & 26.90 & 27.10 \\
4 & 31.74 & 14.99 & 20.36 & 26.41 & 26.17 & 26.29 \\
2 & 29.95 & 12.77 & 17.91 & 23.85 & 25.73 & 24.75 \\
\hline
\end{tabular}
}}}
\caption{\textbf{Influence of number of frames}. Trained on Ego4D. }
\label{tab:num_frames_supmat}
\vspace{-3mm}
\end{table}

%% file: tables/temporal_att.tex
\begin{table}
\centering
\scalebox{1}{ 
\centering
\setlength{\tabcolsep}{2mm}{
\resizebox{0.9\columnwidth}{!}{
\begin{tabular}{c|ccc||ccc} 
\hline
 & \multicolumn{3}{c||}{Nouns} & \multicolumn{3}{c}{Verbs} \\
 \hline
 & \multicolumn{1}{c}{E4D} & \multicolumn{1}{c}{EK} & \multirow{1}{*}{hm} & \multicolumn{1}{c}{E4D} & \multicolumn{1}{c}{EK} & \multirow{1}{*}{hm} \\ 
\hline
w/o & 31.41 & 13.31 & 18.69 & \multicolumn{1}{l}{22.65} & \multicolumn{1}{l}{20.19} & \multicolumn{1}{l}{21.34} \\
w/ & 33.54 & 15.35 & \textbf{21.06} & \multicolumn{1}{l}{28.93} & \multicolumn{1}{l}{26.48} & \multicolumn{1}{l}{\textbf{27.65}} \\
\hline
\end{tabular}
}}}
\caption{\textbf{Influence of temporal attention.} Replacing the temporal module with a simple average decreases verb and noun recognition. The models share the same architecture for Ego-Spatial attention module.}
\label{tab:temp_attention}
\end{table}